\documentclass[final]{cvpr}

\usepackage{times}
\usepackage{epsfig}
\usepackage{graphicx}
\usepackage{amsmath}
\usepackage{amssymb}
\usepackage{xcolor}

\usepackage{algorithm}
\usepackage{listings}

\usepackage[british,american]{babel}

\usepackage{enumitem}

\newcommand{\app}{\raise.17ex\hbox{$\scriptstyle\sim$}}

\newcommand{\dist}{\mathcal{D}}
\newcommand{\p}{{p}}  
\newcommand{\z}{{z}}  

\usepackage{pifont}
\newcommand{\cmark}{\checkmark} 

\newcommand{\appdx}{supplement}

\usepackage{nicefrac,xfrac}

\usepackage{tabulary}
\newcolumntype{x}[1]{>{\centering\arraybackslash}p{#1pt}}

\usepackage{etoolbox}
\makeatletter
\AfterEndEnvironment{algorithm}{\let\@algcomment\relax}
\AtEndEnvironment{algorithm}{\kern2pt\hrule\relax\vskip3pt\@algcomment}
\let\@algcomment\relax
\newcommand\algcomment[1]{\def\@algcomment{\footnotesize#1}}
\renewcommand\fs@ruled{\def\@fs@cfont{\bfseries}\let\@fs@capt\floatc@ruled
  \def\@fs@pre{\hrule height.8pt depth0pt \kern2pt}%
  \def\@fs@post{}%
  \def\@fs@mid{\kern2pt\hrule\kern2pt}%
  \let\@fs@iftopcapt\iftrue}
\makeatother

\definecolor{citecolor}{HTML}{0071bc}
\usepackage[pagebackref=false,breaklinks=true,letterpaper=true,colorlinks,citecolor=citecolor,bookmarks=false]{hyperref}

\newlength\savewidth\newcommand\shline{\noalign{\global\savewidth\arrayrulewidth
  \global\arrayrulewidth 1pt}\hline\noalign{\global\arrayrulewidth\savewidth}}
\newcommand{\tablestyle}[2]{\setlength{\tabcolsep}{#1}\renewcommand{\arraystretch}{#2}\centering\footnotesize}
\makeatletter\renewcommand\paragraph{\@startsection{paragraph}{4}{\z@}
  {.5em \@plus1ex \@minus.2ex}{-.5em}{\normalfont\normalsize\bfseries}}\makeatother

\newcommand{\deh}[1]{\textcolor{gray}{#1}}

\def\cvprPaperID{****} 
\def\confYear{CVPR 2021}

\begin{document}

\title{\vspace{-1.em} Exploring Simple Siamese Representation Learning \vspace{-.5em}}

\author{
Xinlei Chen \qquad Kaiming He \vspace{.5em} \\
Facebook AI Research (FAIR)
\vspace{1.em}  
}

\maketitle

\begin{abstract}
\vspace{-.5em}
Siamese networks have become a common structure in various recent models for unsupervised visual representation learning.
These models maximize the similarity between two augmentations of one image, subject to certain conditions for avoiding collapsing solutions. In this paper, we report surprising empirical results that \mbox{\textbf{simple Siamese}} networks can learn meaningful representations even using \textbf{none} of the following: (i) negative sample pairs, (ii) large batches, (iii) momentum encoders. Our experiments show that collapsing solutions do exist for the loss and structure, but a stop-gradient operation plays an essential role in preventing collapsing.
We provide a hypothesis on the implication of stop-gradient, and further show proof-of-concept experiments verifying it.
Our ``SimSiam'' method achieves competitive results on ImageNet and downstream tasks. 
We hope this simple baseline will motivate people to rethink the roles of Siamese architectures for unsupervised representation learning. Code will be made available.
\vspace{-.5em}
\end{abstract}

\section{Introduction\label{sec:intro}}

Recently there has been steady progress in un-/self-supervised representation learning, with encouraging results on multiple visual tasks (\eg, \cite{Bachman2019,He2019a,Chen2020,Grill2020,Caron2020}). Despite various original motivations, these methods generally involve certain forms of Siamese networks \cite{Bromley1994}. Siamese networks are weight-sharing neural networks applied on two or more inputs. They are natural tools for \emph{comparing} (including but not limited to ``\emph{contrasting}") entities. Recent methods define the inputs as two augmentations of one image, and maximize the similarity subject to different conditions.

\begin{figure}[t]
\centering
\vspace{-1em}
\includegraphics[width=.72\linewidth]{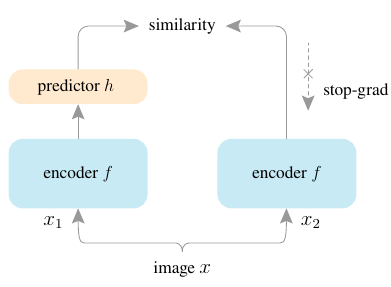}
\vspace{-.5em}
\caption{
\textbf{SimSiam architecture}. Two augmented views of one image are processed by the same encoder network $f$ (a backbone plus a projection MLP). Then a prediction MLP $h$ is applied on one side, and a stop-gradient operation is applied on the other side. The model maximizes the similarity between both sides.
It uses neither negative pairs nor a momentum encoder.
\label{fig:teaser}
\vspace{-.5em}
}
\end{figure}

An undesired trivial solution to Siamese networks is all outputs ``collapsing'' to a constant. 
There have been several general strategies for preventing Siamese networks from collapsing.
Contrastive learning \cite{Hadsell2006}, \eg, instantiated in SimCLR \cite{Chen2020}, repulses different images (negative pairs) while attracting the same image's two views (positive pairs).
The negative pairs preclude constant outputs from the solution space. 
Clustering \cite{Caron2018} is another way of avoiding constant output, and SwAV \cite{Caron2020} incorporates online clustering into Siamese networks.
Beyond contrastive learning and clustering, BYOL \cite{Grill2020} relies only on positive pairs but it does not collapse in case a momentum encoder is used.

In this paper, we report that simple Siamese networks can work surprisingly well with \emph{none} of the above strategies for preventing collapsing. 
Our model directly maximizes the similarity of one image's two views, using \emph{neither} negative pairs \emph{nor} a momentum encoder. It works with typical batch sizes and does not rely on large-batch training.
We illustrate this ``SimSiam'' method in Figure~\ref{fig:teaser}.

Thanks to the conceptual simplicity, SimSiam can serve as a hub that relates several existing methods.
In a nutshell, our method can be thought of as ``BYOL \emph{without} the momentum encoder". Unlike BYOL but like SimCLR and SwAV, our method directly shares the weights between the two branches, so it can also be thought of as ``SimCLR \emph{\mbox{without}} negative pairs", and ``SwAV \emph{without} online clustering". Interestingly, SimSiam is related to each method by removing one of its core components. Even so, SimSiam does not cause collapsing and can perform competitively.

We empirically show that collapsing solutions \emph{do} exist, but a stop-gradient operation (Figure~\ref{fig:teaser}) is critical to prevent such solutions. The importance of stop-gradient suggests that there should be a different underlying optimization problem that is being solved. We hypothesize that there are implicitly two sets of variables, and SimSiam behaves like alternating between optimizing each set.
We provide proof-of-concept experiments to verify this hypothesis.

Our simple baseline suggests that the Siamese architectures can be an essential reason for the common success of the related methods. 
Siamese networks can naturally introduce inductive biases for modeling invariance, as by definition ``invariance'' means that two observations of the same concept should produce the same outputs.
Analogous to convolutions \cite{LeCun1989}, which is a successful inductive bias via weight-sharing for modeling translation-invariance, the weight-sharing Siamese networks can model invariance \wrt more complicated transformations (\eg, augmentations). We hope our exploration will motivate people to rethink the fundamental roles of Siamese architectures for unsupervised representation learning.

\section{Related Work\label{sec:related}}

\paragraph{Siamese networks.} Siamese networks \cite{Bromley1994} are general models for comparing entities.
Their applications include signature \cite{Bromley1994} and face \cite{Taigman2014} verification, tracking \cite{Bertinetto2016}, one-shot learning \cite{Koch2015}, and others. In conventional use cases, the inputs to Siamese networks are from different images, and the comparability is determined by supervision.

\paragraph{Contrastive learning.}
The core idea of contrastive learning \cite{Hadsell2006} is to attract the positive sample pairs and repulse the negative sample pairs. This methodology has been recently popularized for un-/self-supervised representation learning \cite{Wu2018a,Oord2018,Hjelm2019,Ye2019, Henaff2019,Bachman2019,Tian2019,He2019a,Misra2019,Chen2020,Chen2020a}.
Simple and effective instantiations of contrastive learning have been developed using Siamese networks \cite{Ye2019,Bachman2019,He2019a,Chen2020,Chen2020a}.

In practice, contrastive learning methods benefit from a large number of negative samples \cite{Wu2018a,Tian2019,He2019a,Chen2020}. 
These samples can be maintained in a memory bank \cite{Wu2018a}. In a Siamese network, MoCo \cite{He2019a} maintains a queue of negative samples and turns one branch into a momentum encoder to improve consistency of the queue.
SimCLR \cite{Chen2020} directly uses negative samples coexisting in the current batch, and it requires a large batch size to work well.

\paragraph{Clustering.} Another category of methods for unsupervised representation learning are based on clustering \cite{Caron2018,Caron2019,Asano2019,Caron2020}.
They alternate between clustering the representations and learning to predict the cluster assignment.
SwAV \cite{Caron2020} incorporates clustering into a Siamese network, by computing the assignment from one view and predicting it from another view.
SwAV performs online clustering under a balanced partition constraint for each batch, which is solved by the Sinkhorn-Knopp transform \cite{Cuturi2013}.

While clustering-based methods do not define negative \mbox{exemplars}, the cluster centers can play as negative prototypes.
Like contrastive learning, clustering-based methods require either a memory bank \cite{Caron2018,Caron2019,Asano2019}, large batches \cite{Caron2020}, or a queue \cite{Caron2020} to provide enough samples for clustering.

\paragraph{BYOL.} BYOL \cite{Grill2020} directly predicts the output of one view from another view. It is a Siamese network in which one branch is a momentum encoder.\footnote{MoCo \cite{He2019a} and BYOL \cite{Grill2020} do not directly share the weights between the two branches, though in theory the momentum encoder should converge to the same status as the trainable encoder. We view these models as Siamese networks with ``indirect" weight-sharing.}
It is hypothesized in \cite{Grill2020} that the momentum encoder is important for BYOL to avoid collapsing, and it reports failure results if removing the momentum encoder (0.3\% accuracy, Table 5 in \cite{Grill2020}).\footnote{
In BYOL's arXiv v3 update, it reports 66.9\% accuracy with 300-epoch pre-training when removing the momentum encoder and increasing the predictor's learning rate by 10$\times$.
Our work was done concurrently with this arXiv update.
Our work studies this topic from different perspectives, with better results achieved.}
Our empirical study challenges the \emph{necessity} of the momentum encoder for preventing collapsing.
We discover that the stop-gradient operation is critical. This discovery can be obscured with the usage of a momentum encoder, which is always accompanied with stop-gradient (as it is not updated by its parameters' gradients). While the moving-average behavior may improve accuracy with an appropriate momentum coefficient, our experiments show that it is not directly related to preventing collapsing.

\section{Method\label{sec:method}}

\begin{algorithm}[t]
\caption{SimSiam Pseudocode, PyTorch-like}
\label{alg:code}
\definecolor{codeblue}{rgb}{0.25,0.5,0.5}
\definecolor{codekw}{rgb}{0.85, 0.18, 0.50}
\lstset{
  backgroundcolor=\color{white},
  basicstyle=\fontsize{7.5pt}{7.5pt}\ttfamily\selectfont,
  columns=fullflexible,
  breaklines=true,
  captionpos=b,
  commentstyle=\fontsize{7.5pt}{7.5pt}\color{codeblue},
  keywordstyle=\fontsize{7.5pt}{7.5pt}\color{codekw},
}
\begin{lstlisting}[language=python]
# f: backbone + projection mlp
# h: prediction mlp

for x in loader:  # load a minibatch x with n samples
    x1, x2 = aug(x), aug(x)  # random augmentation
    z1, z2 = f(x1), f(x2)  # projections, n-by-d
    p1, p2 = h(z1), h(z2)  # predictions, n-by-d

    L = D(p1, z2)/2 + D(p2, z1)/2   # loss

    L.backward()  # back-propagate
    update(f, h)  # SGD update

def D(p, z):  # negative cosine similarity
    z = z.detach()  # stop gradient

    p = normalize(p, dim=1)  # l2-normalize
    z = normalize(z, dim=1)  # l2-normalize
    return -(p*z).sum(dim=1).mean()
\end{lstlisting}
\end{algorithm}

\begin{figure*}[t]
\vspace{-.5em}
\begin{minipage}[c]{0.8\linewidth}
\includegraphics[width=.33\linewidth]{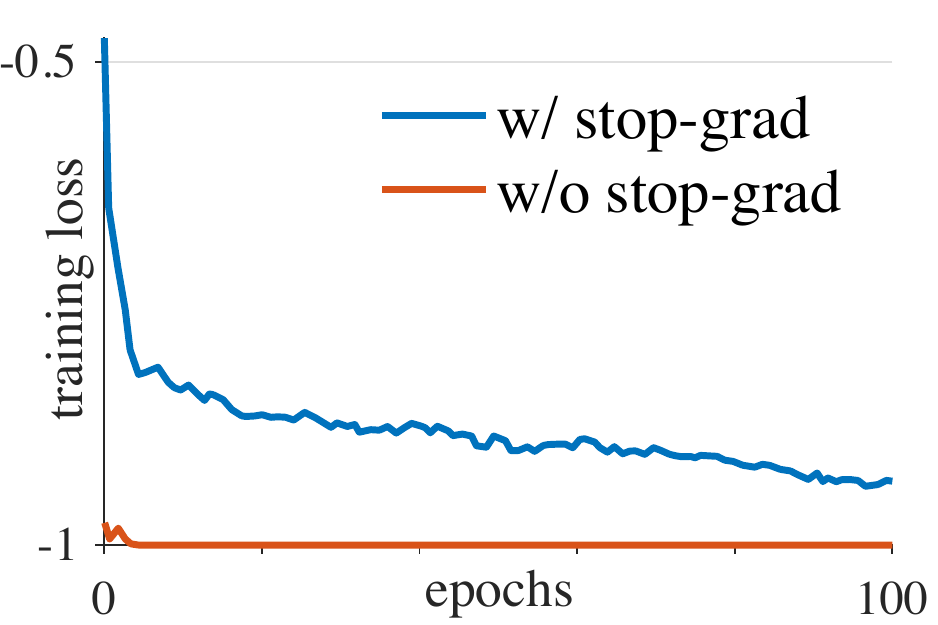}
\includegraphics[width=.33\linewidth]{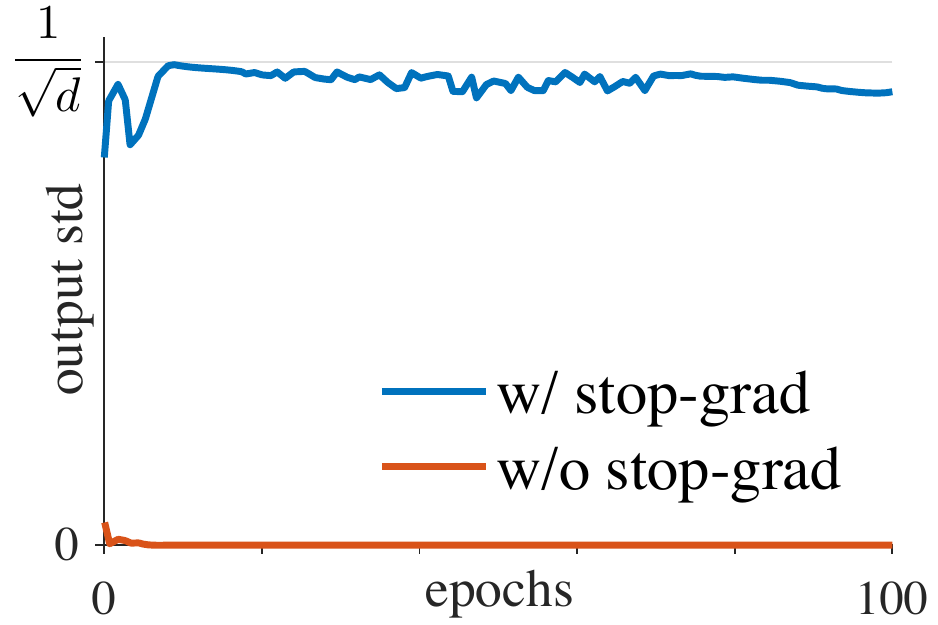}
\includegraphics[width=.33\linewidth]{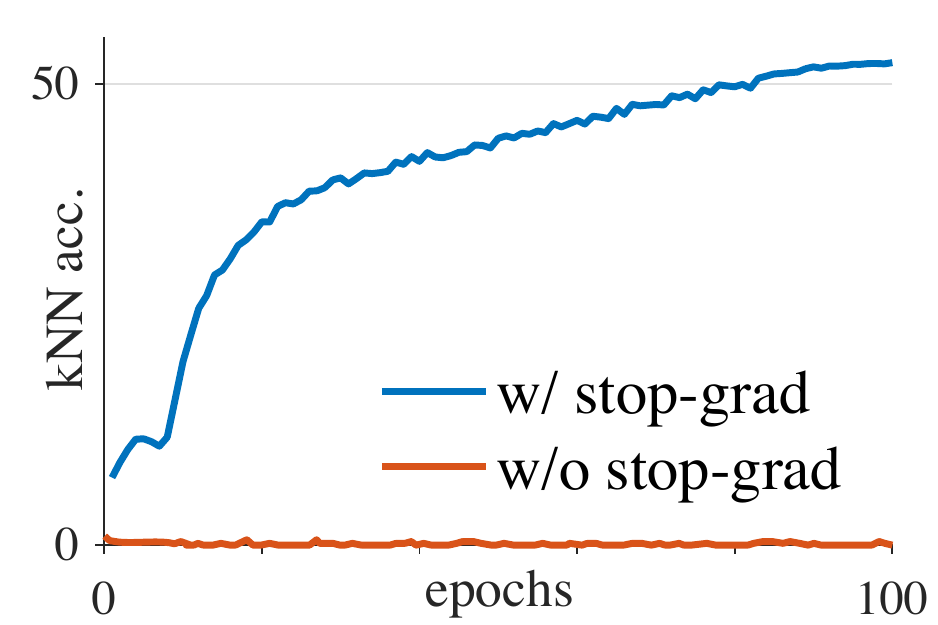}
\end{minipage}
\begin{minipage}[c]{0.19\linewidth}
\small
\tablestyle{4pt}{1.2}
\begin{tabular}{r|c}
 & acc. (\%) \\
\shline
w/ stop-grad & 67.7$\pm$0.1 \\
w/o stop-grad & 0.1 \\
\end{tabular}
\vspace{-.5em}
\end{minipage}
\vspace{.8em}
\caption{\textbf{SimSiam with \vs without stop-gradient}.
\textbf{Left plot}: training loss. Without stop-gradient it degenerates immediately. 
\textbf{Middle plot}: the per-channel std of the $\ell_2$-normalized output, plotted as the averaged std over all channels.
\textbf{Right plot}: validation accuracy of a kNN classifier \cite{Wu2018a} as a monitor of progress.
\textbf{Table}: ImageNet linear evaluation (``w/ stop-grad'' is mean$\pm$std over 5 trials).
\label{fig:stopgrad}
}
\vspace{-.5em}
\end{figure*}

Our architecture (Figure~\ref{fig:teaser}) takes as input two randomly augmented views $x_1$ and $x_2$ from an image $x$.
The two views are processed by an encoder network $f$ consisting of a backbone (\eg, ResNet \cite{He2016}) and a projection MLP head \cite{Chen2020}. The encoder $f$ shares weights between the two views. A prediction MLP head \cite{Grill2020}, denoted as $h$, transforms the output of one view and matches it to the other view.
Denoting the two output vectors as $\p_1\!\triangleq\!h(f(x_1))$ and $\z_2\!\triangleq\!f(x_2)$,
we minimize their negative cosine similarity:
\newcommand{\lnorm}[1]{\frac{#1}{\left\lVert{#1}\right\rVert _2}}
\newcommand{\lnormv}[1]{{#1}/{\left\lVert{#1}\right\rVert _2}}
\begin{equation}
\dist(\p_1, \z_2) = - \lnorm{\p_1}{\cdot}\lnorm{\z_2},
\label{eq:dist_cosine}
\end{equation}
where ${\left\lVert{\cdot}\right\rVert _2}$ is $\ell_2$-norm. This is equivalent to the mean squared error of $\ell_2$-normalized vectors \cite{Grill2020}, up to a scale of 2.
Following \cite{Grill2020}, we define a symmetrized loss as:
\begin{equation}
\mathcal{L} = \frac{1}{2}\dist(\p_1, \z_2) + \frac{1}{2}\dist(\p_2, \z_1).
\label{eq:loss_sym}
\end{equation} 
This is defined for each image, and the total loss is averaged over all images. Its minimum possible value is $-1$.

An important component for our method to work is a \mbox{stop-gradient} ($\texttt{stopgrad}$) operation (Figure~\ref{fig:teaser}). We implement it by modifying (\ref{eq:dist_cosine}) as:
\begin{equation}
\dist(\p_1, \texttt{stopgrad}(\z_2)).
\label{eq:loss_asym_stopgrad}
\end{equation}
This means that $\z_2$ is treated as a constant in this term.
Similarly, the form in (\ref{eq:loss_sym}) is implemented as:
\begin{equation}
\mathcal{L}{=}\frac{1}{2}\dist(\p_1, \texttt{stopgrad}(\z_2)){+}\frac{1}{2}\dist(\p_2, \texttt{stopgrad}(\z_1)).
\label{eq:loss_sym_stopgrad}
\end{equation}
Here the encoder on $x_2$ receives no gradient from $\z_2$ in the first term, but it receives gradients from $\p_2$ in the second term (and vice versa for $x_1$).

The pseudo-code of SimSiam is in Algorithm~\ref{alg:code}.

\paragraph{Baseline settings.} Unless specified, our explorations use the following settings for unsupervised pre-training:
\begin{itemize}[leftmargin=*]

\item \emph{Optimizer}. We use SGD for pre-training. Our method does \emph{not} require a large-batch optimizer such as LARS \cite{You2017} (unlike \cite{Chen2020,Grill2020,Caron2020}).
We use a learning rate of $lr{\times}$BatchSize${/}256$ (linear scaling \cite{Goyal2017}), with a base $lr\!=\!0.05$. The learning rate has a cosine decay schedule \cite{Loshchilov2016,Chen2020}. The weight decay is $0.0001$ and the SGD momentum is $0.9$.

The batch size is 512 by default, which is friendly to typical 8-GPU implementations. Other batch sizes also work well (Sec.~\ref{subsec:batch}). We use batch normalization (BN) \cite{Ioffe2015} synchronized across devices, following \cite{Chen2020,Grill2020,Caron2020}.

\item \emph{Projection MLP}. The projection MLP (in $f$) has BN applied to each fully-connected (fc) layer, including its output fc. Its output fc has no ReLU. The hidden fc is $2048$-d. This MLP has 3 layers.

\item \emph{Prediction MLP}. The prediction MLP ($h$) has BN applied to its hidden fc layers. Its output fc does not have BN (ablation in Sec.~\ref{subsec:bn}) or ReLU. This MLP has 2 layers.
The dimension of $h$'s input and output ($\z$ and $\p$) is $d\!=\!2048$, and $h$'s hidden layer's dimension is $512$, making $h$ a bottleneck structure (ablation in \appdx). 

\end{itemize}

\noindent We use ResNet-50 \cite{He2016} as the default backbone. Other implementation details are in \appdx. We perform 100-epoch pre-training in ablation experiments. 

\paragraph{Experimental setup.}
We do unsupervised pre-training on the 1000-class ImageNet training set \cite{Deng2009} without using labels. The quality of the pre-trained representations is evaluated by training a supervised linear classifier on frozen representations in the training set, and then testing it in the validation set, which is a common protocol. The implementation details of linear classification are in \appdx.

\section{Empirical Study\label{sec:exp}}

In this section we empirically study the SimSiam behaviors. We pay special attention to what may contribute to the model's non-collapsing solutions. 

\subsection{Stop-gradient}\label{subsec:stopgrad}

Figure~\ref{fig:stopgrad} presents a comparison on ``with \vs without stop-gradient". The architectures and all hyper-parameters are kept unchanged, and stop-gradient is the only difference.

Figure~\ref{fig:stopgrad} (left) shows the training loss. \emph{Without} stop-gradient, the optimizer quickly finds a degenerated solution and reaches the minimum possible loss of $-1$.
To show that the degeneration is caused by collapsing, we study the standard deviation (std) of the $\ell_2$-normalized output $\lnormv{\z}$. If the outputs collapse to a constant vector, their std over all samples should be zero for each channel.
This can be observed from the red curve in Figure~\ref{fig:stopgrad} (middle). 

As a comparison, if the output $\z$ has a zero-mean isotropic Gaussian distribution, we can show that the std of $\lnormv{\z}$ is $\frac{1}{\sqrt{d}}$.\footnote{
Here is an informal derivation:
denote $\lnormv{\z}$ as $\z'$, that is, $\z'_{i}\!=\!\z_{i} / (\sum_{j=1}^{d}\z^2_{j})^\frac{1}{2}$ for the $i$-th channel.
If $\z_{j}$ is subject to an i.i.d Gaussian distribution: $\z_{j}\!\sim\!\mathcal{N}(0, 1)$, $\forall j$, then $\z'_{i}\!\approx\!\z_{i} / d^\frac{1}{2}$ and $\text{{std}}[\z'_{i}]\!\approx\!1/d^\frac{1}{2}$.
} 
The blue curve in Figure~\ref{fig:stopgrad} (middle) shows that with stop-gradient, the std value is near $\frac{1}{\sqrt{d}}$. This indicates that the outputs do not collapse, and they are scattered on the unit hypersphere.

Figure~\ref{fig:stopgrad} (right) plots the validation accuracy of a k-nearest-neighbor (kNN) classifier \cite{Wu2018a}. This kNN classifier can serve as a monitor of the progress. With stop-gradient, the kNN monitor shows a steadily improving accuracy.

The linear evaluation result is in the table in Figure~\ref{fig:stopgrad}. SimSiam achieves a nontrivial accuracy of 67.7\%. This result is reasonably stable as shown by the std of 5 trials.
Solely removing stop-gradient, the accuracy becomes 0.1\%, which is the chance-level guess in ImageNet.

\paragraph{Discussion.}
Our experiments show that \emph{there exist collapsing solutions}. 
The collapse can be observed by the minimum possible loss and the constant outputs.\footnote{We note that a chance-level accuracy (0.1\%) is not sufficient to indicate collapsing. A model with a diverging loss, which is another pattern of failure, may also exhibit a chance-level accuracy. }
The existence of the collapsing solutions implies that it is \emph{insufficient} for our method to prevent collapsing \emph{solely} by the architecture designs (\eg, predictor, BN, $\ell_2$-norm). In our comparison, all these architecture designs are kept unchanged, but they do not prevent collapsing if stop-gradient is removed.

The introduction of stop-gradient implies that there should be another optimization problem that is being solved underlying. We propose a hypothesis in Sec.~\ref{sec:hypo}.

\subsection{Predictor} \label{subsec:mlp2}

\begin{table}[t]
\centering
\small
\tablestyle{8pt}{1.1}
\begin{tabular}{cl|r}
& {pred. MLP $h$}  & acc. (\%) \\
\shline
baseline & $lr$ with cosine decay & 67.7 \\
\textbf{(a)} & no pred. MLP & 0.1 \\
\textbf{(b)} & fixed random init. & 1.5 \\
\textbf{(c)} & $lr$ not decayed & 68.1  \\
\end{tabular}
\vspace{.5em}
\caption{\textbf{Effect of prediction MLP}  (ImageNet linear evaluation accuracy with 100-epoch pre-training). In all these variants, we use the same schedule for the encoder $f$ (\mbox{$lr$} with cosine decay). 
\label{tab:mlp2}
}
\end{table}

In Table~\ref{tab:mlp2} we study the predictor MLP's effect. 

The model does not work if removing $h$ (Table~\ref{tab:mlp2}a), \ie, $h$ is the identity mapping.
Actually, this observation can be expected if the symmetric loss (\ref{eq:loss_sym_stopgrad}) is used. Now the loss is $\frac{1}{2}\dist(\z_1, \texttt{stopgrad}(\z_2))$ ${+}$ $\frac{1}{2}\dist(\z_2, \texttt{stopgrad}(\z_1))$. Its gradient has the same direction as the gradient of $\dist(\z_1, \z_2)$, with the magnitude scaled by $1/2$. In this case, using stop-gradient is equivalent to removing stop-gradient and scaling the loss by 1/2. Collapsing is observed (Table~\ref{tab:mlp2}a).

We note that this derivation on the gradient direction is valid only for the symmetrized loss. But we have observed that the \emph{asymmetric} variant (\ref{eq:loss_asym_stopgrad}) also fails if removing $h$,  while it can work if $h$ is kept (Sec.~\ref{subsec:symm}).
These experiments suggest that $h$ is helpful for our model.

If $h$ is fixed as random initialization, our model does not work either (Table~\ref{tab:mlp2}b).
However, this failure is \emph{not} about collapsing. 
The training does not converge, and the loss remains high.
The predictor $h$ should be trained to adapt to the representations. 

We also find that $h$ with a constant $lr$ (without decay) can work well and produce even better results than the baseline (Table~\ref{tab:mlp2}c). A possible explanation is that $h$ should adapt to the latest representations, so it is not necessary to force it converge (by reducing $lr$) before the representations are sufficiently trained. In many variants of our model, we have observed that $h$ with a constant $lr$ provides slightly better results. We use this form in the following subsections.

\subsection{Batch Size} \label{subsec:batch}
 
\begin{table}[t]
\centering
\small
\tablestyle{6pt}{1.1}
\begin{tabular}{c|ccccccc}
batch size & 64 & 128 & 256 & 512 & 1024 & 2048 & 4096 \\
\shline
acc. (\%) & 66.1 & 67.3 & 68.1 & 68.1 & 68.0 & 67.9 & 64.0 \\ 
\end{tabular}
\vspace{.5em}
\caption{\textbf{Effect of batch sizes} (ImageNet linear evaluation accuracy with 100-epoch pre-training).
\label{tab:batch}
}
\end{table}

Table~\ref{tab:batch} reports the results with a batch size from 64 to 4096. When the batch size changes, we use the same linear scaling rule ($lr$$\times$BatchSize/256) \cite{Goyal2017} with base $lr\!=\!0.05$.
We use 10 epochs of warm-up \cite{Goyal2017} for batch sizes $\geq\!$~1024. Note that we keep using the same SGD optimizer (rather than LARS \cite{You2017}) for all batch sizes studied.

Our method works reasonably well over this wide range of batch sizes. Even a batch size of 128 or 64 performs decently, with a drop of 0.8\% or 2.0\% in accuracy.
The results are similarly good when the batch size is from 256 to 2048, and the differences are at the level of random variations.

This behavior of SimSiam is noticeably different from SimCLR \cite{Chen2020} and SwAV \cite{Caron2020}. All three methods are Siamese networks with direct weight-sharing,
but SimCLR and SwAV both require a large batch (\eg, 4096) to work well.

We also note that the standard SGD optimizer does not work well when the batch is too large (even in supervised learning \cite{Goyal2017,You2017}), and our result is lower with a 4096 batch.
We expect a specialized optimizer (\eg, LARS \cite{You2017}) will help in this case. However, our results show that a specialized optimizer is \emph{not} necessary for preventing collapsing.

\subsection{Batch Normalization} \label{subsec:bn}

\begin{table}[t]
\centering
\small
\tablestyle{5pt}{1.1}
\begin{tabular}{cl| cc | cc | c }
~ & ~ & \multicolumn{2}{c|}{proj. MLP's BN} & \multicolumn{2}{c|}{pred. MLP's BN}  & ~ \\
\multicolumn{2}{c|}{case}  & hidden & output & hidden & output & acc. (\%) \\
\shline
\textbf{(a)} & none & - & - & - & - & 34.6 \\
\textbf{(b)} & hidden-only & \cmark & - & \cmark & - & 67.4 \\
\textbf{(c)} & default & \cmark & \cmark & \cmark & - & 68.1 \\
\textbf{(d)} & all & \cmark & \cmark & \cmark & \cmark & unstable \\
\end{tabular}
\vspace{.5em}
\caption{\textbf{Effect of batch normalization on MLP heads} (ImageNet linear evaluation accuracy with 100-epoch pre-training).
\label{tab:bn}
}
\end{table}

Table~\ref{tab:bn} compares the configurations of BN on the MLP heads.
In Table~\ref{tab:bn}a we remove all BN layers in the MLP heads (10-epoch warmup \cite{Goyal2017} is used specifically for this entry).
This variant does \emph{not} cause collapse, although the accuracy is low (34.6\%). The low accuracy is likely because of optimization difficulty.
Adding BN to the hidden layers (Table~\ref{tab:bn}b) increases accuracy to 67.4\%.

Further adding BN to the output of the \emph{projection} MLP (\ie, the output of $f$) boosts accuracy to 68.1\% (Table~\ref{tab:bn}c), which is our default configuration. 
In this entry, we also find that the learnable affine transformation (scale and offset \cite{Ioffe2015}) in $f$'s output BN is not necessary, and disabling it leads to a comparable accuracy of 68.2\%.

Adding BN to the output of the prediction MLP $h$ does not work well (Table~\ref{tab:bn}d). We find that this is not about collapsing. The training is unstable and the loss oscillates.  

In summary, we observe that BN is helpful for optimization when used appropriately, which is similar to BN's behavior in other \emph{supervised} learning scenarios. But we have seen no evidence that BN helps to prevent collapsing: actually, the comparison in Sec.~\ref{subsec:stopgrad} (Figure~\ref{fig:stopgrad}) has exactly the same BN configuration for both entries, but the model collapses if stop-gradient is not used.

\subsection{Similarity Function} \label{subsec:similarity}

Besides the cosine similarity function (\ref{eq:dist_cosine}), our method also works with cross-entropy similarity.
We modify $\dist$ as: $\dist(\p_1, \z_2)\!=\!{-}\texttt{softmax}({\z_2}){\cdot} \log \texttt{softmax}({\p_1})$. Here the softmax function is along the channel dimension. The output of softmax can be thought of as the probabilities of belonging to each of $d$ pseudo-categories.

We simply replace the cosine similarity with the cross-entropy similarity, and symmetrize it using (\ref{eq:loss_sym_stopgrad}).
All hyper-parameters and architectures are unchanged, though they may be suboptimal for this variant. Here is the comparison:
\begin{center}
\vspace{-.2em}
\small
\tablestyle{2pt}{1.1}
\begin{tabular}{x{48} | x{48} x{48}}
& cosine & cross-entropy \\
\shline
acc. (\%) & 68.1 & 63.2 \\
\end{tabular}
\vspace{-.2em}
\end{center}
The cross-entropy variant can converge to a reasonable result without collapsing. This suggests that the collapsing prevention behavior is not just about the cosine similarity.

This variant helps to set up a connection to SwAV \cite{Caron2020}, which we discuss in Sec.~\ref{sec:methodology}.

\subsection{Symmetrization} \label{subsec:symm}

Thus far our experiments have been based on the symmetrized loss (\ref{eq:loss_sym_stopgrad}). We observe that SimSiam's behavior of preventing collapsing does not depend on symmetrization.
We compare with the \emph{asymmetric} variant (\ref{eq:loss_asym_stopgrad}) as follows:
\begin{center}
\vspace{-.2em}
\small
\tablestyle{2pt}{1.1}
\begin{tabular}{x{48} | x{48}x{48}x{48}}
 & sym. & asym. & asym. 2$\times$ \\
\shline
\multicolumn{1}{c|}{acc. (\%)} & 68.1 & 64.8 & 67.3  \\
\end{tabular}
\vspace{-.2em}
\end{center}
The asymmetric variant achieves reasonable results.
Symmetrization is helpful for boosting accuracy, but it is not related to collapse prevention. Symmetrization makes one more prediction for each image, and we may roughly compensate for this by sampling two pairs for each image in the asymmetric version (``2$\times$''). It makes the gap smaller.

\subsection{Summary}

We have empirically shown that in a variety of settings, SimSiam can produce meaningful results without collapsing. The optimizer (batch size), batch normalization, similarity function, and symmetrization may affect accuracy, but we have seen \emph{no} evidence that they are related to collapse prevention. It is mainly the stop-gradient operation that plays an essential role.

\section{Hypothesis} \label{sec:hypo}

We discuss a hypothesis on what is implicitly optimized by SimSiam, with proof-of-concept experiments provided.

\subsection{Formulation}

Our hypothesis is that SimSiam is an implementation of an Expectation-Maximization (EM) like algorithm. It implicitly involves two sets of variables, and solves two underlying sub-problems. The presence of stop-gradient is the consequence of introducing the extra set of variables.

We consider a loss function of the following form:
\begin{equation}
\mathcal{L}(\theta, \eta)=
\mathbb{E}_{x, \mathcal{T}}
\Big[
	\big\|
	\mathcal{F}_\theta(\mathcal{T}(x)) - \eta_{x} 
	\big\|_2^2
\Big].
\label{eq:em_loss}
\end{equation}
$\mathcal{F}$ is a network parameterized by $\theta$.
$\mathcal{T}$ is the augmentation.
$x$ is an image.
The expectation $\mathbb{E}[\cdot]$ is over the distribution of images and augmentations.
For the ease of analysis, here we use the mean squared error $\|\cdot\|^2_2$, which is equivalent to the cosine similarity if the vectors are $\ell_2$-normalized.
We do not consider the predictor yet and will discuss it later.

In (\ref{eq:em_loss}), we have introduced \emph{another set of variables} which we denote as $\eta$.
The size of $\eta$ is proportional to the number of images.
Intuitively, $\eta_x$ is the representation of the image $x$, and the subscript $_x$ means using the image index to access a sub-vector of $\eta$. 
$\eta$ is \emph{not necessarily} the output of a network; it is the argument of an optimization problem.

With this formulation, we consider solving:
\begin{equation}
\min_{\theta, \eta} \mathcal{L}(\theta, \eta).
\label{eq:em_problem}
\end{equation}
Here the problem is \wrt both $\theta$ and $\eta$.
This formulation is analogous to k-means clustering \cite{MacQueen1967}. The variable $\theta$ is analogous to the clustering centers: it is the learnable parameters of an encoder. The variable $\eta_x$ is analogous to the assignment vector of the sample $x$ (a one-hot vector in k-means): it is the representation of $x$. 

Also analogous to k-means, the problem in (\ref{eq:em_problem}) can be solved by an alternating algorithm, fixing one set of variables and solving for the other set. Formally, we can alternate between solving these two subproblems:
\begin{eqnarray}
\theta^t &\leftarrow & \arg\min_{\theta}~\mathcal{L}(\theta, \eta^{t{-}1})
\label{eq:em_subproblem1}
\\
\eta^t &\leftarrow& \arg\min_{\eta}~\mathcal{L}(\theta^t, \eta)
\label{eq:em_subproblem2}
\end{eqnarray}
Here $t$ is the index of alternation and ``$\leftarrow$" means assigning.

\paragraph{Solving for $\theta$.} One can use SGD to solve the sub-problem (\ref{eq:em_subproblem1}).
\emph{The stop-gradient operation is a natural consequence}, because the gradient does not back-propagate to $\eta^{t-1}$ which is a constant in this subproblem.

\paragraph{Solving for $\eta$.} The sub-problem (\ref{eq:em_subproblem2}) can be solved independently for each $\eta_x$. Now the problem is to minimize:
$
\mathbb{E}_{\mathcal{T}}
\Big[
	\|
	\mathcal{F}_{\theta^t}(\mathcal{T}(x)) - \eta_{x} 
	\|^2_2
\Big]
$
for each image $x$, noting that the expectation is over the distribution of augmentation $\mathcal{T}$.
Due to the mean squared error,\footnote{If we use the cosine similarity, we can approximately 
solve it by $\ell_2$-normalizing $\mathcal{F}$'s output and $\eta_x$.} it is easy to solve it by:
\begin{equation}
\eta^t_x \leftarrow
\mathbb{E}_{\mathcal{T}}
\Big[
	\mathcal{F}_{\theta^t}(\mathcal{T}(x))
\Big].
\label{eq:eta0}
\end{equation}
This indicates that $\eta_x$ is assigned with the \emph{average} representation of $x$ over the distribution of augmentation.

\paragraph{One-step alternation.} SimSiam can be approximated by one-step alternation between (\ref{eq:em_subproblem1}) and (\ref{eq:em_subproblem2}).
First, we approximate (\ref{eq:eta0}) by 
sampling the augmentation only \emph{once}, denoted as $\mathcal{T}'$, and ignoring $\mathbb{E}_{\mathcal{T}}[\cdot]$:
\begin{equation}
\eta^{t}_x \leftarrow
	\mathcal{F}_{\theta^t}(\mathcal{T'}(x)).
\label{eq:eta}
\end{equation}
Inserting it into the sub-problem (\ref{eq:em_subproblem1}), we have:
\begin{gather}
\theta^{t+1} \leftarrow \arg\min_{\theta}
\mathbb{E}_{x, \mathcal{T}}
\Big[
	\big\|
	\mathcal{F}_\theta(\mathcal{T}(x)) - \mathcal{F}_{\theta^{t}}(\mathcal{T'}(x))
	\big\|_2^2
\Big].
\label{eq:em_simsiam}
\end{gather}
Now $\theta^{t}$ is a constant in this sub-problem, and $\mathcal{T'}$ implies another view due to its random nature. This formulation exhibits the Siamese architecture. 
Second, if we implement (\ref{eq:em_simsiam}) by \emph{reducing} the loss with one SGD step, then we can approach the SimSiam algorithm: a Siamese network naturally with stop-gradient applied.

\paragraph{Predictor.}
Our above analysis does not involve the predictor $h$. We further assume that $h$ is helpful in our method because of the approximation due to (\ref{eq:eta}).

By definition, the predictor $h$ is expected to minimize:
$
\mathbb{E}_z
\Big[
	\big\|
	h(\z_1)
	-
	z_2
	\big\|^2_2
\Big]
$.
The optimal solution to $h$ should satisfy:
$
h(z_1)\!=\!\mathbb{E}_z
[
	z_2
]
\!=\!
\mathbb{E}_{\mathcal{T}}
\big[
	f(\mathcal{T}(x))
\big]
$ for any image $x$.
This term is similar to the one in (\ref{eq:eta0}).
In our approximation in (\ref{eq:eta}),
the expectation $\mathbb{E}_{\mathcal{T}}[\cdot]$ is ignored. The usage of $h$ may fill this gap.
In practice, it would be unrealistic to actually compute the expectation $\mathbb{E}_{\mathcal{T}}$. But it may be possible for a neural network (\eg, the preditor $h$) to learn to predict the expectation, while the sampling of $\mathcal{T}$ is implicitly distributed across multiple epochs. 

\paragraph{Symmetrization.} Our hypothesis does not involve symmetrization. Symmetrization is like denser sampling $\mathcal{T}$ in (\ref{eq:em_simsiam}). Actually, the SGD optimizer computes the empirical expectation of $\mathbb{E}_{x, \mathcal{T}}[\cdot]$ by sampling a batch of images and one pair of augmentations ($\mathcal{T}_1$, $\mathcal{T}_2$). In principle, the empirical expectation should be more precise with denser sampling. Symmetrization supplies an extra pair ($\mathcal{T}_2$, $\mathcal{T}_1$).
This explains that symmetrization is not necessary for our method to work, yet it is able to improve accuracy, as we have observed in Sec.~\ref{subsec:symm}.

\subsection{Proof of concept} \label{subsec:poc}

We design a series of proof-of-concept experiments that stem from our hypothesis. They are methods different with SimSiam, and they are designed to verify our hypothesis.

\paragraph{Multi-step alternation.}
We have hypothesized that the 
SimSiam algorithm is like alternating between (\ref{eq:em_subproblem1}) and (\ref{eq:em_subproblem2}), with an interval of one step of SGD update. Under this hypothesis, it is likely for our formulation to work if the interval has multiple steps of SGD. 

In this variant, we treat $t$ in (\ref{eq:em_subproblem1}) and (\ref{eq:em_subproblem2}) as the index of an outer loop; and the sub-problem in (\ref{eq:em_subproblem1}) is updated by an inner loop of $k$ SGD steps. 
In each alternation, we \mbox{pre-compute} the $\eta_x$ required for all $k$ SGD steps using (\ref{eq:eta}) and cache them in memory.
Then we perform $k$ SGD steps to update $\theta$.
We use the same architecture and hyper-parameters as SimSiam.
The comparison is as follows:
{
\begin{center}
\vspace{-.2em}
\small
\tablestyle{1pt}{1.1}
\begin{tabular}{x{42} | x{42} x{42} x{42} x{42}}
& 1-step & 10-step & 100-step & 1-epoch \\
\shline
acc. (\%) & 68.1 & 68.7 & 68.9 & 67.0 \\
\end{tabular}
\vspace{-.2em}
\end{center}
}
\noindent Here, ``1-step" is equivalent to SimSiam, and ``1-epoch'' denotes the $k$ steps required for one epoch. All multi-step variants work well. The 10-/100-step variants even achieve \emph{better} results than SimSiam, though at the cost of extra pre-computation.
This experiment suggests that the alternating optimization is a valid formulation, and SimSiam is a special case of it.

\begin{table*}[t]
\vspace{-1em}
\centering
\small
\tablestyle{2pt}{1.1}
\begin{tabular}{l x{36}x{36}x{36}|x{28}x{28}x{28}x{28}}
method
& {\tablestyle{0pt}{.9} \begin{tabular}{c} {batch} \\ {size} \end{tabular}}
& {\tablestyle{0pt}{.9} \begin{tabular}{c} {negative} \\ {pairs} \end{tabular}}
& {\tablestyle{0pt}{.9} \begin{tabular}{c} {momentum} \\ {encoder} \end{tabular}}
& 100 ep & 200 ep & 400 ep & 800 ep \\
\shline
SimCLR (repro.+) & 4096 & \cmark & &
66.5 & 68.3 & 69.8 & 70.4 \\
MoCo v2 (repro.+) & \textbf{256} & \cmark & \cmark &
67.4 & 69.9 & 71.0 & 72.2 \\
BYOL (repro.) & 4096 & & \cmark &
66.5 & \textbf{70.6} & \textbf{73.2} & \textbf{74.3} \\
SwAV (repro.+) & 4096 & & &
66.5 & 69.1 & 70.7 & 71.8 \\
\hline
\textbf{SimSiam} & \textbf{256} & & & 
\textbf{68.1} & 70.0 & 70.8 & 71.3 \\
\end{tabular}
\vspace{.5em}
\caption{
\textbf{Comparisons on ImageNet linear classification}. All are based on \textbf{ResNet-50} pre-trained with \textbf{two 224$\times$224 views}. Evaluation is on a single crop.  All competitors are from our reproduction, and ``+'' denotes \emph{improved} reproduction \vs original papers (see \appdx).\label{tab:sota}
}
\end{table*}

\begin{table*}[t]
\centering
\small
\tablestyle{2pt}{1.1}
\begin{tabular}{l | x{24}x{24}x{24} | x{24}x{24}x{24} | x{24}x{24}x{24} | x{24}x{24}x{24} }
& \multicolumn{3}{c|}{VOC 07 detection} & \multicolumn{3}{c|}{VOC 07+12 detection} & \multicolumn{3}{c|}{COCO detection} & \multicolumn{3}{c}{COCO instance seg.} \\
pre-train
& AP$_\text{50}$ & AP & AP$_\text{75}$
& AP$_\text{50}$ & AP & AP$_\text{75}$
& AP$_\text{50}$ & AP & AP$_\text{75}$
& AP$^\text{mask}_\text{50}$ & AP$^\text{mask}$ & AP$^\text{mask}_\text{75}$ \\
\shline
\deh{scratch} & \deh{35.9} & \deh{16.8} & \deh{13.0} & \deh{60.2} & \deh{33.8} & \deh{33.1} & \deh{44.0} & \deh{26.4} & \deh{27.8} & \deh{46.9} & \deh{29.3} & \deh{30.8} \\
ImageNet supervised & 74.4 & 42.4 & 42.7 & 81.3 & 53.5 & 58.8 & 58.2 & 38.2 & 41.2 & 54.7 & 33.3 & 35.2
\\
\hline
SimCLR (repro.+) & 75.9 & 46.8 & 50.1 & 81.8 & 55.5 & 61.4 & 57.7 & 37.9 & 40.9 & 54.6 & 33.3 & 35.3 \\
MoCo v2 (repro.+) & \textbf{77.1} & \textbf{48.5} & \textbf{52.5} & \textbf{82.3} & \textbf{57.0} & \textbf{63.3} & \textbf{58.8} & \textbf{39.2} & \textbf{42.5} & \textbf{55.5} & \textbf{34.3} & \textbf{36.6} \\
BYOL (repro.) & \textbf{77.1} & 47.0 & 49.9 & 81.4 & 55.3 & 61.1 & 57.8 & 37.9 & 40.9 & 54.3 & 33.2 & 35.0 \\
SwAV (repro.+) & 75.5 & 46.5 & 49.6 & 81.5 & 55.4 & 61.4 & 57.6 & 37.6 & 40.3 & 54.2 & 33.1 & 35.1 \\
\hline
\textbf{SimSiam}, base & 75.5 & 47.0 & 50.2 & \textbf{82.0} & 56.4 & 62.8 & 57.5 & 37.9 & 40.9 & 54.2 & 33.2 & 35.2 \\
 \textbf{SimSiam}, optimal & \textbf{77.3} & \textbf{48.5} & \textbf{52.5} & \textbf{82.4} & \textbf{57.0} & \textbf{63.7} & \textbf{59.3} & \textbf{39.2} & \textbf{42.1} & \textbf{56.0} & \textbf{34.4} & \textbf{36.7} \\
\end{tabular}
\vspace{.5em}
\caption{\textbf{Transfer Learning}.
All unsupervised methods are based on 200-epoch pre-training in ImageNet. \emph{VOC 07 detection}: Faster R-CNN \cite{Ren2015} fine-tuned in VOC 2007 trainval, evaluated in VOC 2007 test; \emph{VOC 07+12 detection}: Faster R-CNN fine-tuned in VOC 2007 trainval + 2012 train, evaluated in VOC 2007 test; \emph{COCO detection} and \emph{COCO instance segmentation}:  Mask R-CNN \cite{He2017} (1$\times$ schedule) fine-tuned in
COCO 2017 train, evaluated in COCO 2017 val.
All Faster/Mask R-CNN models are with the C4-backbone \cite{Detectron2018}.
All VOC results are the average over 5 trials. 
\textbf{Bold entries} are within 0.5 below the best.
\label{tab:transfer}
}
\vspace{-1em}
\end{table*}

\paragraph{Expectation over augmentations.} The usage of the predictor $h$ is presumably because the expectation $\mathbb{E}_{\mathcal{T}}[\cdot]$ in (\ref{eq:eta0}) is ignored. We consider another way to approximate this expectation, in which we find $h$ is not needed.

In this variant, we do not update $\eta_x$ directly by the assignment (\ref{eq:eta}); instead, we maintain a moving-average: $\eta^t_x \leftarrow m* \eta^{t-1}_x + (1-m)* \mathcal{F}_{\theta^t}(\mathcal{T'}(x))$, where $m$ is a momentum coefficient (0.8 here). This computation is similar to maintaining the \emph{memory bank} as in \cite{Wu2018a}.
This moving-average provides an approximated expectation of multiple views.
This variant has 55.0\% accuracy \emph{without} the predictor $h$. As a comparison, it fails completely if we remove $h$ but do not maintain the moving average (as shown in Table~\ref{tab:mlp2}a). This proof-of-concept experiment supports that the usage of predictor $h$ is related to approximating $\mathbb{E}_{\mathcal{T}}[\cdot]$.

\subsection{Discussion}

Our hypothesis is about what the optimization problem can be. It does not explain why collapsing is prevented.
We point out that SimSiam and its variants' non-collapsing behavior still remains as an empirical observation.

Here we briefly discuss our understanding on this open question. 
The alternating optimization provides a different trajectory, and the trajectory depends on the initialization.
It is \emph{unlikely} that the initialized $\eta$, which is the output of a randomly initialized network, would be a constant.
Starting from this initialization, it may be difficult for the alternating optimizer to approach a constant $\eta_x$ for all $x$, because the method does \emph{not} compute the gradients \wrt $\eta$ jointly for all $x$.
The optimizer seeks another trajectory (Figure~\ref{fig:stopgrad} left), in which the outputs are scattered (Figure~\ref{fig:stopgrad} middle).

\section{Comparisons} \label{sec:results}

\subsection{Result Comparisons} \label{sec:results}

\paragraph{ImageNet.}
We compare with the state-of-the-art frameworks in Table~\ref{tab:sota} on ImageNet linear evaluation. For fair comparisons, all competitors are based on our reproduction, and ``+'' denotes \emph{improved} reproduction \vs the original papers (see \appdx).
For each individual method, we follow the hyper-parameter and augmentation recipes in its original paper.\footnote{In our BYOL reproduction, the 100, 200(400), 800-epoch recipes follow the 100, 300, 1000-epoch recipes in \cite{Grill2020}: $lr$ is \{0.45, 0.3, 0.2\}, $wd$ is \{1e-6, 1e-6, 1.5e-6\}, and momentum coefficient is \{0.99, 0.99, 0.996\}.}
All entries are based on a standard ResNet-50, with two 224$\times$224 views used during pre-training.

Table~\ref{tab:sota} shows the results and the main properties of the methods.
SimSiam is trained with a batch size of 256, using neither negative samples nor a momentum encoder.
Despite it simplicity, SimSiam achieves competitive results.
It has the highest accuracy among all methods under 100-epoch pre-training, though its gain of training longer is smaller. 
It has better results than SimCLR in all cases.

\paragraph{Transfer Learning.} In Table~\ref{tab:transfer} we compare the representation quality by transferring them to other tasks, including VOC \cite{Everingham2010} object detection and COCO \cite{Lin2014} object detection and instance segmentation.
 We fine-tune the pre-trained models end-to-end in the target datasets. We use the public codebase from MoCo \cite{He2019a} for all entries, and search the fine-tuning learning rate for each individual method.
All methods are based on 200-epoch pre-training in ImageNet using our reproduction.

Table~\ref{tab:transfer} shows that SimSiam's representations are \emph{transferable} beyond the ImageNet task.
It is competitive among these leading methods. The ``base'' SimSiam in Table~\ref{tab:transfer} uses the baseline pre-training recipe as in our ImageNet experiments. We find that another recipe of $lr\!=\!\text{0.5}$ and $wd\!=\!\text{1e-5}$ (with similar ImageNet accuracy) can produce better results in all tasks (Table~\ref{tab:transfer}, ``SimSiam, optimal").

We emphasize that \emph{all these methods are highly successful for transfer learning}---in Table~\ref{tab:transfer}, they can surpass or be on par with the ImageNet \emph{supervised} pre-training counterparts in all tasks.
Despite many design differences, a \mbox{common} structure of these methods is the \emph{Siamese network}.
This comparison suggests that the Siamese structure is a core factor for their general success.

\subsection{Methodology Comparisons} \label{sec:methodology}

\begin{figure}[t]
\begin{center}
\includegraphics[width=.99\linewidth]{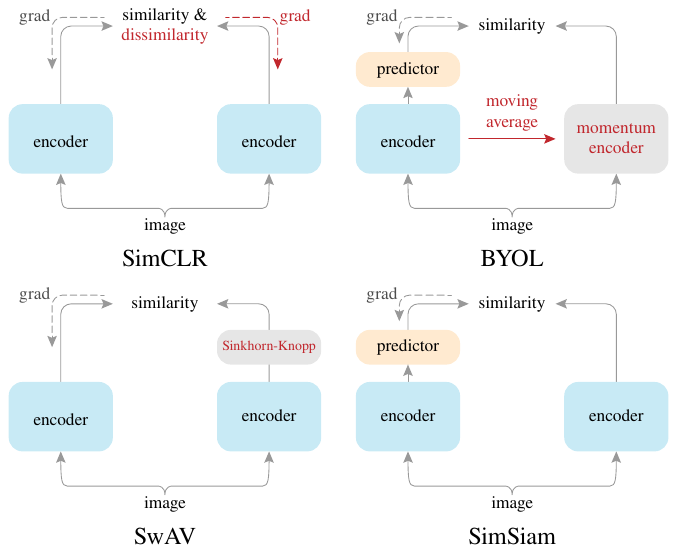}
\end{center}
\vspace{-1em}
\caption{
\textbf{Comparison on Siamese architectures}.
The encoder includes all layers that can be shared between both branches. 
The dash lines indicate the gradient propagation flow.
In BYOL, SwAV, and SimSiam, the lack of a dash line implies stop-gradient, and their symmetrization is not illustrated for simplicity. The components in red are those missing in SimSiam.
\label{fig:methodology}
}
\end{figure}

Beyond accuracy, we also compare the methodologies of these Siamese architectures. 
Our method plays as a hub to connect these methods.
Figure~\ref{fig:methodology} abstracts these methods. The ``encoder'' subsumes all layers that can be shared between both branches (\eg, backbone, projection MLP \cite{Chen2020}, prototypes \cite{Caron2020}). The components in red are those missing in SimSiam. We discuss the relations next.

\paragraph{Relation to SimCLR} \cite{Chen2020}. 
SimCLR relies on negative samples (``dissimilarity'') to prevent collapsing.
SimSiam can be thought of as ``SimCLR without negatives".

To have a more thorough comparison, we append the prediction MLP $h$ and stop-gradient to SimCLR.\footnote{We append the extra predictor to one branch and stop-gradient to the other branch, and symmetrize this by swapping.}
Here is the ablation on our SimCLR reproduction:
{
\begin{center}
\vspace{-.2em}
\small
\tablestyle{1pt}{1.1}
\begin{tabular}{x{72} x{72} x{72}}
SimCLR & w/ predictor & w/ pred. \& stop-grad \\
\shline
66.5 & 66.4 & 66.0 \\
\end{tabular}
\vspace{-.2em}
\end{center}
}
\noindent Neither the stop-gradient nor the extra predictor is necessary or helpful for SimCLR. 
As we have analyzed in Sec.~\ref{sec:hypo}, the introduction of the stop-gradient and extra predictor is presumably a consequence of another underlying optimization problem.
It is different from the contrastive learning problem, so these extra components may not be helpful.

\paragraph{Relation to SwAV} \cite{Caron2020}. SimSiam is conceptually \mbox{analogous} to ``SwAV without online clustering".
We build up this connection by recasting a few components in SwAV.
(i) The shared prototype layer in SwAV can be \mbox{absorbed} into the Siamese encoder.
(ii) The prototypes were \mbox{weight-normalized} outside of gradient propagation in \cite{Caron2020};
we instead implement by full gradient computation \cite{Salimans2016}.\footnote{This modification produces similar results as original SwAV, but it can enable end-to-end propagation in our ablation.}
(iii) The similarity function in SwAV is cross-entropy.
With these abstractions, a highly simplified SwAV illustration is shown in Figure~\ref{fig:methodology}. 

SwAV applies the Sinkhorn-Knopp (SK) transform \cite{Cuturi2013} on the target branch (which is also symmetrized \cite{Caron2020}). 
The SK transform is derived from online clustering \cite{Caron2020}: it is the outcome of clustering the current batch subject to a balanced partition constraint. The balanced partition can avoid collapsing.
Our method does not involve this transform.

We study the effect of the prediction MLP $h$ and stop-gradient on SwAV. 
Note that SwAV applies stop-gradient on the SK transform, so we ablate by removing it.
Here is the comparison on our SwAV reproduction:
{
\begin{center}
\vspace{-.2em}
\small
\tablestyle{1pt}{1.1}
\begin{tabular}{x{72} x{72} x{72}}
SwAV & w/ predictor & remove stop-grad \\
\shline
66.5 & 65.2 & NaN \\
\end{tabular}
\vspace{-.2em}
\end{center}
}
\noindent Adding the predictor does not help either. 
Removing stop-gradient (so the model is trained end-to-end) leads to divergence.
As a clustering-based method, SwAV is inherently an alternating formulation \cite{Caron2020}. This may explain why stop-gradient should not be removed from SwAV.

\paragraph{Relation to BYOL} \cite{Grill2020}. Our method can be thought of as ``BYOL without the momentum encoder", subject to many implementation differences. The momentum encoder may be beneficial for accuracy (Table~\ref{tab:sota}), but it is not necessary for preventing collapsing.
Given our hypothesis in Sec.~\ref{sec:hypo}, the $\eta$ sub-problem (\ref{eq:em_subproblem2}) can be solved by other optimizers, \eg, a gradient-based one. This may lead to a temporally smoother update on $\eta$. 
Although not directly related, the momentum encoder also produces a smoother version of $\eta$. We believe that other optimizers for solving (\ref{eq:em_subproblem2}) are also plausible, which can be a future research problem.

\section{Conclusion\label{sec:conc}}

We have explored Siamese networks with simple designs. The competitiveness of our minimalist method suggests that the Siamese shape of the recent methods can be a core reason for their effectiveness. Siamese networks are natural and effective tools for modeling invariance, which is a focus of representation learning. We hope our study will attract the community's attention to the fundamental role of Siamese networks in representation learning.

{\small
\definecolor{Gray}{gray}{0.5}
\renewcommand\UrlFont{\color{Gray}\rmfamily}
\bibliographystyle{ieee_fullname}
\bibliography{simsiam}
}

\newpage

\renewcommand\thefigure{\thesection.\arabic{figure}}
\renewcommand\thetable{\thesection.\arabic{table}}
\setcounter{figure}{0} 
\setcounter{table}{0} 

\appendix

\section{Implementation Details}

\paragraph{Unsupervised pre-training.} Our implementation follows the practice of existing works \cite{Wu2018a,He2019a,Chen2020,Chen2020a,Grill2020}.

\emph{Data augmentation}. We describe data augmentation using the PyTorch \cite{Paszke2019} notations. Geometric augmentation is \texttt{RandomResizedCrop} with scale in $[0.2, 1.0]$ \cite{Wu2018a} and \texttt{RandomHorizontalFlip}. Color augmentation is \texttt{ColorJitter} with \{brightness, contrast, saturation, hue\} strength of \{0.4, 0.4, 0.4, 0.1\} with an applying probability of 0.8, and \texttt{RandomGrayscale} with an applying probability of 0.2. Blurring augmentation \cite{Chen2020} has a Gaussian kernel with std in $[0.1, 2.0]$.

\emph{Initialization}. The convolution and fc layers follow the default PyTorch initializers. Note that by default PyTorch initializes fc layers' weight and bias by a uniform distribution $\mathcal{U}(-\sqrt{k}, \sqrt{k})$ where $k{=}\frac{1}{\text{in\_channels}}$. Models with substantially different fc initializers (\eg, a fixed std of 0.01) may not converge. Moreover, similar to the implementation of \cite{Chen2020}, we initialize the scale parameters as 0 \cite{Goyal2017} in the last BN layer for every residual block.

\emph{Weight decay}. We use a weight decay of 0.0001 for all parameter layers, including the BN scales and biases, in the SGD optimizer. This is in contrast to the implementation of \cite{Chen2020,Grill2020} that excludes BN scales and biases from weight decay in their LARS optimizer.

\paragraph{Linear evaluation.} Given the pre-trained network, we train a supervised linear classifier on frozen features, which are from ResNet's global average pooling layer (pool$_5$). 
The linear classifier training uses base $lr\!=\!0.02$ with a cosine decay schedule for 90 epochs, weight decay$~\!=\!0$, momentum$~\!=\!0.9$, batch size$~\!=\!4096$ with a LARS optimizer \cite{You2017}. We have also tried the SGD optimizer following \cite{He2019a} with base $lr\!=\!30.0$, weight decay$~\!=\!0$, momentum$~\!=\!0.9$, and batch size$~\!=\!256$, which gives \app1\% lower accuracy.
After training the linear classifier, we evaluate it on the center $224{\times}224$ crop in the validation set.

\section{Additional Ablations on ImageNet} \label{subsec:sensitivity}

The following table reports the SimSiam results \vs the output dimension $d$:
\begin{center}
\vspace{-.2em}
\small
\tablestyle{1pt}{1.1}
\begin{tabular}{x{48} | x{28}x{28}x{28}x{28}}
output $d$ & 256 & 512 & 1024 & 2048 \\
\shline
acc. (\%) & 65.3 & 67.2 & 67.5 & 68.1 \\
\end{tabular}
\vspace{-.2em}
\end{center}
It benefits from a larger $d$ and gets saturated at $d\!=\!\text{2048}$. This is unlike existing methods \cite{Wu2018a,He2019a,Chen2020,Grill2020} whose accuracy is saturated when $d$ is 256 or 512. 

In this table, the prediction MLP's hidden layer dimension is always 1/4 of the output dimension. We find that this bottleneck structure is more robust. If we set the hidden dimension to be equal to the output dimension, the training can be less stable or fail in some variants of our exploration.
We hypothesize that this bottleneck structure, which behaves like an auto-encoder, can force the predictor to digest the information.
We recommend to use this bottleneck structure for our method.

\section{Reproducing Related Methods}

Our comparison in Table~\ref{tab:sota} is based on our reproduction of the related methods. We re-implement the related methods as faithfully as possible following each individual paper. In addition, we are able to improve SimCLR, MoCo v2, and SwAV by small and straightforward modifications: specifically, we use 3 layers in the projection MLP in SimCLR and SwAV (\vs originally 2), and use symmetrized loss for MoCo v2 (\vs originally asymmetric). Table~\ref{tab:repro} compares our reproduction of these methods with the original papers' results (if available). Our reproduction has \emph{better} results for SimCLR, MoCo v2, and SwAV (denoted as ``+'' in Table~\ref{tab:sota}), and has at least comparable results for BYOL.

\begin{table}[t]
\centering
\small
\tablestyle{4pt}{1.1}
\begin{tabular}{l | rrr | rr | rrr | c }
& \multicolumn{3}{c|}{SimCLR}
& \multicolumn{2}{c|}{MoCo v2}
& \multicolumn{3}{c|}{BYOL}
& \multicolumn{1}{c}{SwAV}
\\
epoch
& 200 & 800 & 1000  
& 200 & 800  
& 300 & 800 & 1000  
& 400  
\\
\shline
origin
& 66.6 & 68.3 & 69.3  
& 67.5 & 71.1  
& 72.5 & - & 74.3  
& 70.1   
\\
repro.
& 68.3 & 70.4 & -  
& 69.9 & 72.2  
& 72.4 & 74.3 & -  
& 70.7   
\end{tabular}
\vspace{.3em}
\caption{
\textbf{Our reproduction \vs original papers' results}. All are based on ResNet-50 pre-trained with two 224$\times$224 crops.
\label{tab:repro}
\vspace{-.5em}
}
\end{table}

\section{CIFAR Experiments}

We have observed similar behaviors of SimSiam in the CIFAR-10 dataset \cite{Krizhevsky2009}.
The implementation is similar to that in ImageNet.
We use SGD with base $lr\!=\!0.03$ and a cosine decay schedule for 800 epochs, weight decay$~\!=\!0.0005$, momentum$~\!=\!0.9$, and batch size$~\!=\!512$. The input image size is 32$\times$32. We do not use blur augmentation. The backbone is the CIFAR variant of ResNet-18 \cite{He2016}, followed by a 2-layer projection MLP. The outputs are 2048-d.

\begin{figure}[h]
\begin{minipage}[c]{0.64\linewidth}
\includegraphics[width=.99\linewidth]{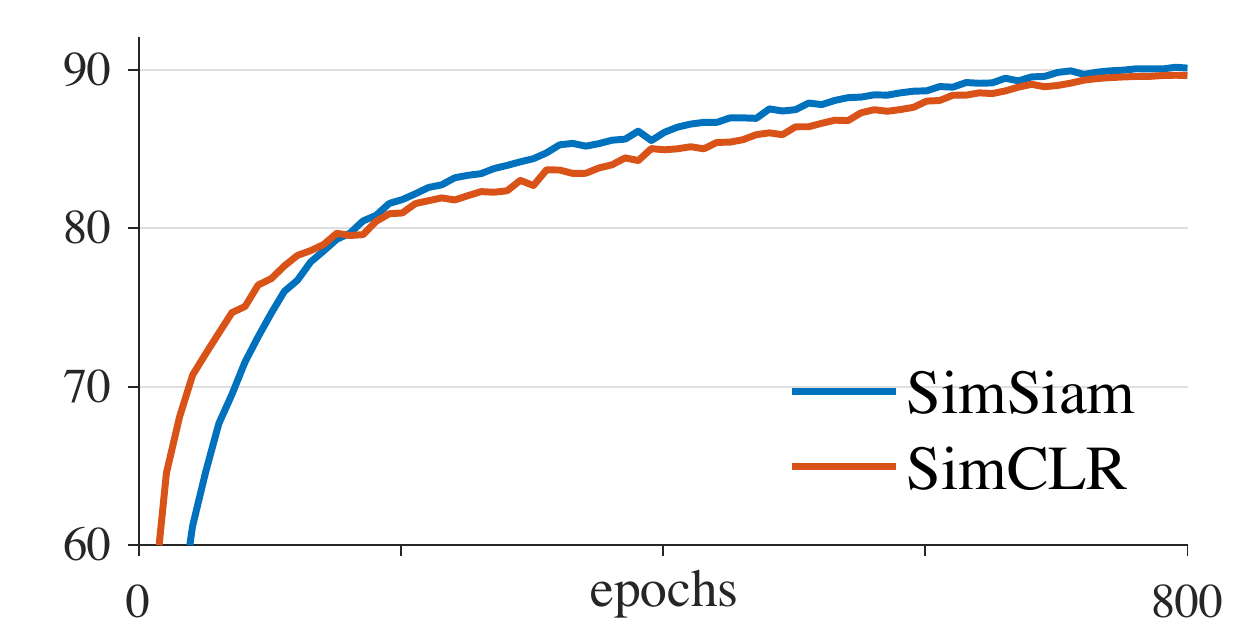}
\end{minipage}
\begin{minipage}[c]{0.19\linewidth}
\small
\tablestyle{4pt}{1.2}
\begin{tabular}{l|c}
 & acc. (\%) \\
\shline
SimCLR & 91.1 \\
SimSiam & 91.8 \\
\end{tabular}
\vspace{-.5em}
\end{minipage}
\vspace{.8em}
\caption{
\textbf{CIFAR-10 experiments}. Left: validation accuracy of kNN classification as a monitor during pre-training. Right: linear evaluation accuracy.
The backbone is ResNet-18.
\label{fig:cifar}
\vspace{-.5em}
}
\end{figure}

Figure~\ref{fig:cifar} shows the kNN classification accuracy (left) and the linear evaluation (right). 
Similar to the ImageNet observations, SimSiam achieves a reasonable result and does not collapse.
We compare with SimCLR \cite{Chen2020} trained with the same setting. Interestingly, the training curves are similar between SimSiam and SimCLR. SimSiam is slightly better by 0.7\% under this setting.

\end{document}